# SINGLE REDUCT GENERATION BASED ON RELATIVE INDISCERNIBILITY OF ROUGH SET THEORY

Shampa Sengupta and Asit Kr. Das

M.C.K.V.Institute Of Engineering - 243, G.T.Road (North), Liluah, Howrah 711204, West Bengal

Bengal Engineering And Science University, Shibpur, Howrah 711103, West Bengal

shampa2512@yahoo.co.in, akdas@cs.becs.ac.in

## Abstract.

*In real world everything is an object which represents particular classes. Every object can be fully described by its attributes. Any real world dataset contains large number of attributes and objects. Classifiers give poor performance when these huge datasets are given as input to it for proper classification. So from these huge dataset most useful attributes need to be extracted that contribute the maximum to the decision. In the paper, attribute set is reduced by generating reducts using the indiscernibility relation of Rough Set Theory (RST). The method measures similarity among the attributes using relative indiscernibility relation and computes attribute similarity set. Then the set is minimized and an attribute similarity table is constructed from which attribute similar to maximum number of attributes is selected so that the resultant minimum set of selected attributes (called reduct) cover all attributes of the attribute similarity table. The method has been applied on glass dataset collected from the UCI repository and the classification accuracy is calculated by various classifiers. The result shows the efficiency of the proposed method.*

## 1. Introduction

In general, considering all attributes highest accuracy of a classifier should be achieved. But for real-world problems, there is huge number of attributes, which degrades the efficiency of the Classification algorithms. So, some attributes need to be neglected, which again decrease the accuracy of the system. Therefore, a trade-off is required for which strong dimensionality reduction or feature selection techniques are needed. The attributes contribute the most to the decision must be retained. Rough Set Theory (RST) [1, 2], new mathematical approach to imperfect knowledge, is popularly used to evaluate significance of attribute and helps to find minimal set of attribute called reduct. Thus a reduct is a set of attributes that preserves partition. It means that a reduct is the minimal subset of attributes that enables the same classification of elements of the universe as the whole set of attributes. In other words, attributes that do not belong to a reduct are superfluous with regard to classification of elements of the universe. Hu et al. [3] developed two new algorithms to calculate core attributes and reducts for feature selection. These algorithms can be extensively applied to a wide range of real-life applications with very large data sets. Jensen et al. [4] developed the Quickreduct algorithm to compute a minimal reduct without exhaustively generating all possible subsets and also they developed Fuzzy-Rough attribute reduction with application to web categorization. Zhong et al. [5] applies Rough Sets with Heuristics (RSH) and Rough Sets with Boolean Reasoning (RSBR) are used for attribute selection and discretization of real-valued attributes. Komorowsk et al. [6] studies an application of rough sets to modelling prognostic power of cardiac tests. Bazan [7] compares rough set-based methods, in particular dynamic reducts, with statistical methods, neural networks, decision trees and decision rules. Carlin et al. [8] presents an application of rough sets to diagnosing suspected acute appendicitis.The main advantage of rough set theory in data analysis is that it does not need any preliminary or additional information about data like probability in statistics [9], or basic probability assignment in Dempster-Shafer theory [10], grade of membership or the value of possibility in fuzzy set theory [11] and so on. But finding reduct for classification is an NP-Complete problem and so some heuristic approach should be applied. In the paper, a novel reduct generation method is proposed based on the indiscernibility relation of rough set theory.

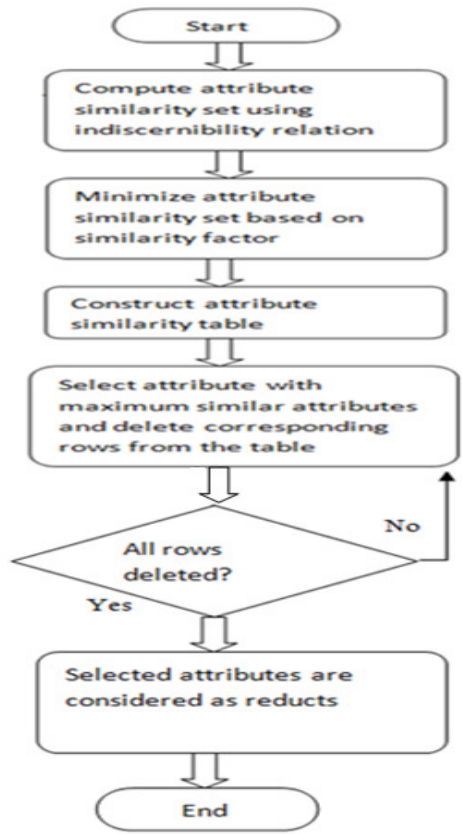

Fig1: Single Reduct Generation Process

In the method, a new kind of indiscernibility, called relative indiscernibility of an attribute with respect to other attribute is introduced. This relative indiscernibility relation induces the partitions of attributes, based on which similarity between conditional attributes is measured and an attribute similarity set (ASS) is obtained. Then, the similarity set is minimized by removing the attribute similarities having similarity measure less than the average similarity. Lastly, an attribute similarity table is constructed for ASS each row of which describes the similarity of an attribute with some other attributes. Then traverse each row and select the attribute of that row which has maximum similar attributes. Next, all the rows associated with the selected attribute and its similar attributes are deleted from the table and similarly select another attribute from the modified table. The process continued until all the rows are deleted from the table and finally, selected attributes, covering all the attributes are considered as reduct, a minimum set of attributes.

The rest of the paper is organized as follows: Similarity measurement of attributes by relative indiscernibility and single reduct generation are described in section 2 and section 3 respectively. Section 4 explains the experimental analysis of the proposed method and finally conclusion of the paper is stated in section 5.

## 2. Relative Indiscernibility and Dependency of Attributes

Formally, a decision system DS can be seen as a system $DS = (U, A)$ where $U$ is the universe (a finite set of objects, $U = <x_1, x_2,...x_m>$) and $A$ is the set of attributes such that $A = C \cup D$ and $C \cap D = \emptyset$ where $C$ and $D$ are the set of condition attributes and the set of decision attributes, respectively.

### 2.1 Indiscernibility

A per the discussion in section II, each attribute $a \in A$ defines an information function: $f_a : U \rightarrow V_a$, where $V_a$ is the set of values of $a$, called the domain of attribute. Every subset of attributes $P$ determines an indiscernibility relation over $U$, and is denoted as $IND(P)$ , which can be defined as, $IND(P) = \{(x, y) \in U \times U \mid \forall a \in P, \ f_a (x) = f_a (y)\}$. For each set of attributes $P$, an indiscernibility relation IND(P) partitions the set of objects into a m-number of equivalence classes [ ] defined as partition U/IND(P) or U/P is equal to $\{[x]_p\}$ where |U/P| = m. Elements belonging to the same equivalence class are indiscernible; otherwise elements are discernible with respect to P. If one considers a non-empty attributes subset, $R \subset P$ and $IND(R) = IND(P)$, then $P − R$ is dispensable. Any minimal R such that $IND(R) = IND(P)$ , is a minimal set of attributes that preserves the indiscernibility relation computed on the set of attributes P. R is called reduct of P and denoted as R = RED(P). The core of P is the intersection of these reductions, defined as $CORE(P) = \cap RED(P)$. Naturally, the core contains all the attributes from P which are considered of greatest importance for classification, i.e., the most relevant for a correct classification of the objects of U. On the other hand, none of the attributes belonging to the core may be neglected without deteriorating the quality of the classification considered, that is, if any attribute in the core is eliminated from the given data, it will be impossible to obtain the highest quality of approximation with the remaining attributes.

### 2.2 Relative Indiscernibility

Here, the relation is defined based on the same information function: $f_a : U \rightarrow V_a$ where $V_a$ is the set of values of a, called the domain of attribute. Every conditional attribute $A_i$ of C determines an relative (relative to decision attribute) indiscernibility relation (RIR) over U, and is denoted as $RIRD(A_i)$, which can be defined by equation (1).

$$RIR_D(A_i) = \{(x,y) \in \Pi_{A_i}[x]_D \times \Pi_{A_i}[x]_D \mid f_{A_i}(x) = f_{A_i}(y) \forall [x]_D \in U/D\} \qquad (1)$$

For each conditional attribute $A_i$, a relative indiscernibility relation $RIR_D(A_i)$ partitions the set of objects into a n-number of equivalence classes [ ] defined as partition $U/ RIR_D(A_i)$ or $U_D/A_i$ is equal to $[[x]_{A_{i,D}}]$ where $\mid U_D/A_i \mid$ = n. Obviously, each equivalence class $[[x]_{A_{i,D}}]$ contains objects with same decision value which are indiscernible by attribute $A_i$.

To illustrate the method, a sample dataset represented by Table 1 is considered with eight objects, four conditional and one decision attributes.

**Table 1.** Sample Dataset

|  | *Diploma*(i) | *Experience*(e) | *French*(f) | *Reference*(r) | *Decision* |
|---|---|---|---|---|---|
| $x_1$ | MBA | Medium | Yes | Excellent | Accept |
| $x_2$ | MBA | Low | Yes | Neutral | Reject |
| $x_3$ | MCE | Low | Yes | Good | Reject |
| $x_4$ | MSc | High | Yes | Neutral | Accept |
| $x_5$ | MSc | Medium | Yes | Neutral | Reject |
| $x_6$ | MSc | High | Yes | Excellent | Reject |
| $x_7$ | MBA | High | No | Good | Accept |
| $x_8$ | MCE | Low | No | Excellent | Reject |

**Table 2**. Equivalence classes induces by indiscernibility and relative indiscernibility relations

| Equivalence classes for each attribute by relation IND(P) | Equivalence classes for each conditional attribute by relative indiscernibility relation $RIRD(A_i)$ |
|---|---|
| $U/_D = (\{x_1, x_4, x_7\}, \{x_2, x_3, x_5, x_6, x_8\})$ | $UD/_i = (\{x_1, x_7\}, \{x_2\}, \{x_3, x_8\}, \{x_4\}, \{x_5, x_6\})$ |
| $U/_i = (\{x_1, x_2, x_7\}, \{x_3, x_8\}, \{x_4, x_5, x_6\})$ | $UD/_e = (\{x_1\}, \{x_5\}, \{x_2, x_3, x_8\}, \{x_4, x_7\}, \{x_6\})$ |
| $U/_e = (\{x_1, x_5\}, \{x_2, x_3, x_8\}, \{x_4, x_6, x_7\})$ | $UD/_f = (\{x_1, x_4\}, \{x_2, x_3, x_5, x_6\}, \{x_7\}, \{x_8\})$ |
| $U/_f = (\{x_1, x_2, x_3, x_4, x_5, x_6\}, \{x_7, x_8\})$ | $UD/_r = (\{x_1\}, \{x_6, x_8\}, \{x_2, x_5\}, \{x_4\}, \{x_3, x_7\})$ |
| $U/_r = (\{x_1, x_6, x_8\}, \{x_2, x_4, x_5\}, \{x_3, x_7\})$ | |

## 2.3 Attribute Similarity

An attribute $A_i$ is similar to another attribute $A_j$ in context of classification power if they induce the same equivalence classes of objects under their respective relative indiscernible relations. But in real situation, it rarely occurs and so similarity of attributes is measured by introducing the similarity measurement factor which indicates the degree of similarity of one attribute to another attribute. Here, an attribute $A_i$ is said to be similar to an attribute $A_j$ with degree of similarity (or similarity factor) $\delta_f^{i,j}$ and is denoted by $A_i \rightarrow A_j$ if the probability of inducing the same equivalence classes of objects under their respective relative indiscernible relations is $(\delta_f^{i,j} \times 100)\%$, where $\delta_f^{i,j}$ is computed by equation (2). The details for computation of similarity measurement for the attribute similarity $A_i \rightarrow A_j$ ($A_i \neq A_j$) is described in algorithm "SIM_FAC" below.

$$\delta_f^{i,j} = \frac{1}{|U_D/A_i|} \sum_{[x]_{A_{i/D}} \in U_D/A_i} \frac{1}{|[x]_{A_{i/D}}|} \max_{[x]_{A_{j/D}} \in U_D/A_j} ([x]_{A_{i/D}} \cap [x]_{A_{j/D}}) \quad (2)$$

Algorithm: SIM_FAC($A_i$ , $A_j$)/* Similarity factor computation for attribute similarity $A_i \rightarrow A_j$ */
Input: Partitions $U_D/A_i = \{[x]_{A_{i/D}}\}$ and $U_D/A_j = \{[x]_{A_{j/D}}\}$

    obtained by applying relative indiscernibility relation $RIR_D$ on

    $A_i$ and $A_j$ respectively.

Output: Similarity factor $\delta_f^{i,j}$

  Begin

    For each conditional attribute $A_i$ {

     /* compute relative indiscernibility RIRD ($A_i$) using (1)*/

      $RIR_D(A_i) = \{(x,y) \in \Pi_{A_i}[x]_D \times \Pi_{A_i}[x]_D \mid f_{A_i}(x) = f_{A_i}(y) \forall [x]_D \in U/D\}$

     $RIR_D$ ($A_i$) induces equivalence classes $U_D/A_i = \{[x]_{A_{i/D}}\}$

    } /*end of for*/

    /* similarity measurement of $A_i$ to $A_j$ */

    $\delta_f^{i,j} = 0$

    For each $[x]_{i/D} \in U_D/A_i$

    { max_overlap = 0

     For each $[x]_{j/D} \in U_D/A_j$

      { overlap = $|[x]_{i/D} \cap [x]_{j/D}|$

       if (overlap > max_overlap) then

        max_overlap = overlap

      }

      $\delta_f^{i,j} = \delta_f^{i,j} + \frac{\text{max\_overlap}}{|[x]_{i/D}|}$

    }

    $\delta_f^{i,j} = \frac{\delta_f^{i,j}}{|U_D/A_i|}$

End.

To illustrate the attribute similarity computation process, attribute similarity and its similarity factor are listed in Table 2 for all attributes of Table 1.

**Table 2.** Describe the degree of similarity of all pair of attributes

| Attribute Similarity<br><br>$(A_i \rightarrow A_j)$ | Equivalence Classes by $RIR_D(A_i)$<br>$(U_D/A_i)$ | Equivalence Classes by $RIR_D(A_j)$<br>$(U_D/A_j)$ | Similarity factor of $A_i$ to $A_j$<br>$(\partial_f^{i,j})$ |
|---|---|---|---|
| $i \rightarrow e$ | $\{x_1, x_7\}$, $\{x_2\}$, $\{x_3, x_8\}$, $\{x_4\}$, $\{x_5, x_6\}$ | $\{x_1\}$, $\{x_5\}$, $\{x_2, x_3, x_8\}$, $\{x_4, x_7\}$, $\{x_6\}$ | $\partial_f^{i,e} = 0.8$ |
| $i \rightarrow f$ | $\{x_1, x_7\}$, $\{x_2\}$, $\{x_3, x_8\}$, $\{x_4\}$, $\{x_5, x_6\}$ | $\{x_1, x_4\}$, $\{x_2, x_3, x_5, x_6\}$, $\{x_7\}$, $\{x_8\}$ | $\partial_f^{i,f} = 0.8$ |
| $i \rightarrow r$ | $\{x_1, x_7\}$, $\{x_2\}$, $\{x_3, x_8\}$, $\{x_4\}$, $\{x_5, x_6\}$ | $\{x_1\}$, $\{x_6, x_8\}$, $\{x_2, x_5\}$, $\{x_4\}$, $\{x_3, x_7\}$ | $\partial_f^{i,r} = 0.7$ |
| $e \rightarrow i$ | $\{x_1\}$, $\{x_5\}$, $\{x_2, x_3, x_8\}$, $\{x_4, x_7\}$, $\{x_6\}$ | $\{x_1, x_7\}$ , $\{x_2\}$, $\{x_3, x_8\}$, $\{x_4\}$, $\{x_5, x_6\}$ | $\partial_f^{e,i} = 0.83$ |
| $e \rightarrow f$ | $\{x_1\}$, $\{x_5\}$, $\{x_2, x_3, x_8\}$, $\{x_4, x_7\}$, $\{x_6\}$ | $\{x_1, x_4\}$, $\{x_2, x_3, x_5, x_6\}$, $\{x_7\}$, $\{x_8\}$ | $\partial_f^{e,f} = 0.83$ |
| $e \rightarrow r$ | $\{x_1\}$, $\{x_5\}$, $\{x_2, x_3, x_8\}$, $\{x_4, x_7\}$, $\{x_6\}$ | $\{x_1\}$, $\{x_6, x_8\}$, $\{x_2, x_5\}$, $\{x_4\}$, $\{x_3, x_7\}$ | $\partial_f^{e,r} = 0.76$ |
| $f \rightarrow i$ | $\{x_1, x_4\}$, $\{x_2, x_3, x_5, x_6\}$, $\{x_7\}$, $\{x_8\}$ | $\{x_1, x_7\}$ , $\{x_2\}$, $\{x_3, x_8\}$, $\{x_4\}$, $\{x_5, x_6\}$ | $\partial_f^{f,i} = 0.75$ |
| $f \rightarrow e$ | $\{x_1, x_4\}$, $\{x_2, x_3, x_5, x_6\}$, $\{x_7\}$, $\{x_8\}$ | $\{x_1\}$, $\{x_5\}$, $\{x_2, x_3, x_8\}$, $\{x_4, x_7\}$, $\{x_6\}$ | $\partial_f^{f,e} = 0.75$ |
| $f \rightarrow r$ | $\{x_1, x_4\}$, $\{x_2, x_3, x_5, x_6\}$, $\{x_7\}$, $\{x_8\}$ | $\{x_1\}$, $\{x_6, x_8\}$, $\{x_2, x_5\}$, $\{x_4\}$, $\{x_3, x_7\}$ | $\partial_f^{f,r} = 0.75$ |
| $r \rightarrow i$ | $\{x_1\}$, $\{x_6, x_8\}$, $\{x_2, x_5\}$, $\{x_4\}$, $\{x_3, x_7\}$ | $\{x_1, x_7\}$ , $\{x_2\}$, $\{x_3, x_8\}$, $\{x_4\}$, $\{x_5, x_6\}$ | $\partial_f^{r,i} = 0.7$ |
| $r \rightarrow e$ | $\{x_1\}$, $\{x_6, x_8\}$, $\{x_2, x_5\}$, $\{x_4\}$, $\{x_3, x_7\}$ | $\{x_1\}$, $\{x_5\}$, $\{x_2, x_3, x_8\}$, $\{x_4, x_7\}$, $\{x_6\}$ | $\partial_f^{r,i} = 0.7$ |
| $r \rightarrow f$ | $\{x_1\}$, $\{x_6, x_8\}$, $\{x_2, x_5\}$, $\{x_4\}$, $\{x_3, x_7\}$ | $\{x_1, x_4\}$, $\{x_2, x_3, x_5, x_6\}$, $\{x_7\}$, $\{x_8\}$ | $\partial_f^{r,f} = 0.8$ |

The computation of $\partial_f^{i,j}$ of each attribute similarity using equation (2) in Table 2 can be understood by Table 3, in which similarity $i \rightarrow e$ in first row of Table 2 is considered, where, $U_D/i = \{x1, x7\}$, $\{x2\}$, $\{x3, x8\}$, $\{x4\}$, $\{x5, x6\}$) and $U_D/e = \{x1\}$, $\{x5\}$, $\{x2, x3, x8\}$, $\{x4, x7\}$, $\{x6\}$).

**Table 3.** Illustrates the similarity factor computation for i → e

| $[x]_{i/D}$ of $U_D/i$ | Overlapping $[x]_{e/D}$ of $U_D/e$ with $[x]_{i/D}$ of $U_D/i$ | $[x]_{i/D} \cap [x]_{e/D}$ | $T = \dfrac{1}{\|[x]_{i/D}\|} \max_{[x]_{e/D} \in U_D/e}(\|[x]_{i/D} \cap [x]_{e/D}\|)$ |
|---|---|---|---|
| $\{x_1, x_7\}$ | $\{x_1\}$<br>$\{x_4, x_7\}$ | $\{x_1, x_7\} \cap \{x_1\}$<br>$\{x_1, x_7\} \cap \{x_4, x_7\}$ | $\dfrac{1}{2}$ |
| $\{x_2\}$ | $\{x_2, x_3, x_8\}$ | $\{x_2\} \cap \{x_2, x_3, x_8\}$ | $\dfrac{1}{1}$ |
| $\{x_3, x_8\}$ | $\{x_2, x_3, x_8\}$ | $\{x_3, x_8\} \cap \{x_2, x_3, x_8\}$ | $\dfrac{2}{2}$ |
| $\{x_4\}$ | $\{x_4, x_7\}$ | $\{x_4\} \cap \{x_4, x_7\}$ | $\dfrac{1}{1}$ |
| $\{x_5, x_6\}$ | $\{x_5\}$<br>$\{x_6\})$ | $\{x_5, x_6\} \cap \{x_5\}$<br>$\{x_5, x_6\} \cap \{x_6\}$ | $\dfrac{1}{2}$ |

$$\delta_f^{ie} = \frac{1}{\|[x]_{i/D}\|} \sum [x]_{i/D} \in U_D/i\, T = \frac{1}{5}\left(\frac{1}{2} + \frac{1}{1} + \frac{2}{2} + \frac{1}{1} + \frac{1}{2}\right) = \frac{4}{5} = 0.8$$

## 2.4 Attribute Similarity Set

For each pair of conditional attributes (Ai, Aj), similarity factor is computed by "SIM_FAC" algorithm, described in section 2.3. The similarity factor of $A_i \rightarrow A_j$ is higher means that the relative indiscernibility relations RIRD(Ai) and RIRD(Aj) produce highly similar equivalence classes. This implies that both the attributes Ai and Aj have almost similar classification power and so $A_i \rightarrow A_j$ is considered as strong similarity of Ai to Aj. Since, for any two attributes Ai and Aj, two similarities $A_i \rightarrow A_j$ and $A_j \rightarrow A_i$ are computed, only one with higher similarity factor is selected in the list of attribute similarity set ASS. Thus, for n conditional attributes, n(n-1)/2similarities are selected, out of which some are strong and some are not. Out of these similarities, the similarity with $\delta_f^{i,j}$ value less than the average $\delta_f$ value are discarded from ASS and rest is considered as the set of attribute similarity. So, each element x in ASS is of the form x: Ai→Aj such that Left(x) = Ai and Right(x) = Aj. The algorithm "ASS_GEN" described below, computes the attribute similarity set ASS.

Algorithm: ASS_GEN(C, $\delta_f$)

/* Computes attribute similarity set {Ai→Aj} */

Input: C = set of conditional attributes and δf =2-D contains    similarity factors between each pair of conditional attributes.

Output: Attribute Similarity Set ASS

Begin

ASS = { }, sum_$\delta_f$ = 0

/* compute only n(n − 1)/2 elements in ASS */

for i = 1 to |C| - 1

{  for j = i+1 to |C|

  {  if($\delta_f^{i,j} > \delta_f^{j,i}$)then

    {   sum_$\delta_f$ = sum_$\delta_f$ + $\delta_f^{i,j}$

      ASS = ASS $\cup$ {Ai $\to$ Aj}

    }

    else

    {   sum_$\delta_f$ = sum_$\delta_f$ + $\delta_f^{j,i}$

      ASS = ASS $\cup$ {Aj $\to$ Ai}

    }

  }

}

/* modify ASS by only elements Ai $\to$ Aj for which $\delta_f^{i,j}$>avg_$\delta_f$ */

ASS$_{mod}$ = { }

    avg_$\delta_f$ = (2$\times$ sum_$\delta_f$) / |C|(|C|-1)

for each {Ai $\to$ Aj}$\in$ ASS

{   if($\delta_f^{i,j}$ >avg_$\delta_f$) then

  {  ASS_mod = ASS$_{mod}$ $\cup$ {Ai $\to$ Aj}

    ASS = ASS − { Ai $\to$ Aj}

  }

}

ASS = ASS$_{mod}$

  End

Algorithm "ASS_GEN" is applied and Table 4 is constructed from Table 2, where only six out of twelve attribute similarities in Table 2 are considered. Thus, initially, ASS = {i $\to$ f, i $\to$ r, e $\to$ i, e $\to$ f, e $\to$ r, r $\to$ f} and avg_$\delta_f$= 0.786.  As the similarity factor for attribute similarities i $\to$ f, e $\to$ i, e $\to$ f and r $\to$ f are greater than avg_$\delta_f$, they are considered in the final attribute similarity set ASS. So, finally, ASS = {i $\to$ f, e $\to$ i, e $\to$ f, r $\to$ f }.

**Table 4.** Illustrates the selection of attribute similarities.

| Attribute Similarity<br><br>( $A_i \rightarrow A_j$; $i \neq j$ and $\delta_F^{i,j} > \delta_F^{j,i}$ ) | Similarity factor of $A_i$ to $A_j$<br><br>($\delta_F^{i,j}$) | $\delta_F^{i,j} > \delta_f$ |
|---|---|---|
| i→f | $\delta_f^{i,f} = 0.8$ | Yes |
| i→r | $\delta_f^{i,r} = 0.7$ | |
| e→i | $\delta_f^{e,i} = 0.83$ | Yes |
| e→f | $\delta_f^{e,f} = 0.83$ | Yes |
| e→r | $\delta_f^{e,r} = 0.76$ | |
| r→f | $\delta_f^{r,f} = 0.8$ | Yes |
| Average $\delta_f$ | 0.786 | |

# 3. Single Reduct Generation

The attribute similarity obtained so far is known as simple similarity of an attribute to other attribute. But, for simplifying the reduct generation process, the elements in ASS are minimized by combining some simple similarity. The new similarity obtained by the combination of some of the simple similarity is called compound similarity. Here, all x from ASS with same Left(x) are considered and obtained compound similarity is Left(x) → ∪ Right(x) ∀x. Thus, introducing compound similarity, the set ASS is refined to a set with minimum elements so that for each attribute, there is at most one element in ASS representing either simple or compound similarity of the attribute. The detail algorithm for determining compound attribute similarity set is given below:

Algorithm: COMP_SIM(ASS)

   /* Compute the compound attribute similarity of attributes*/

   Input: Simple attribute similarity set ASS

   Output: Compound attribute similarity set ASS

   Begin

      for each x ∈ ASS

      {   for each y (≠x) ∈ ASS

         {   if(Left(x) = = Left(y)) then

            {   Right(x) = Right(x) ∪ Right(y)

              ASS = ASS – {y}

           }

        }

}

End

Finally, from the compound attribute similarity set ASS, reduct is generated. First of all, select an element, say, x from ASS for which length of Right(x) i.e., |Right(x)| is maximum. This selection guaranteed that the attribute Left(x) is similar to maximum number of attributes and so Left(x) is an element of reduct RED. Then, all elements z of ASS for which Left(z) $\subseteq$ Right(x) are deleted and also x is deleted from ASS. This process is repeated until the set ASS becomes empty which provides the reduct RED. The proposed single reduct generation algorithm is discussed below:

Algorithm: SIN_RED_GEN(ASS, RED)

  Input: Compound attribute similarity set ASS

  Output: Single reduct RED

  Begin

     RED = $\phi$

     While (ASS $\neq \phi$)

    {   max = 0

       for each x $\in$ ASS

       { if(|Right(x)| > max) then

         {  max = |Right(x)|

           L = Left(x)

         }

       }

      for each x $\in$ ASS

      {  if (Left(x) = = L) then

        {  RED = RED $\cup$ Left(x)

         R = Right(x)

         ASS = ASS $-$ {x}

         for each z $\in$ ASS

            if(Left(z) $\subseteq$ R) then

              ASS = ASS $-$ {z}

          break

        }

      }

    } /*end-while*/

    Return (RED)

End

Applying "COMP_SIM" algorithm the set ASS = {i → f, e → i, e → f, r → f} is refined to compound similarity set ASS = {i → f, e → {i, f}, r → f}. So, the selected element from ASS is e → {i, f}, and thus e ∈ RED and ASS is modified as ASS = {r → f}. And, in the next iteration, r ∈ RED and ASS =φ. Thus, RED = {e, r}.

# 4. Results and discussions

The proposed method computes a single reduct for datasets collected from UCI machine learning repository [12]. At first, all the numeric attributes are discretized by ChiMerge [13] discretization algorithm .To measure the efficiency of the method, k-fold cross-validations, where k ranges from 1 to 10 have been carried out on the dataset and classified using "Weka" tool [14]. The proposed method (PRP) and well known dimensionality reduction methods, such as,Correlated Feature Subset (CFS) method [15] and *Consistency Subset Evaluator* (CON) method [16] have been applied on the dataset for dimension reduction and the reduced datasets are classified on various classifiers. Original number of attributes, number of attributes after applying various reduction methods and the accuracies (in %) of the datasets are computed and listed in Table 5, which shows the efficiency of the proposed method.

**Table 5.** Accuracy Comparison of Proposed, CFS and CON methods

| Class ifier | Machine (7) | | | Heart(13) | | | Wine(13) | | | Liver disorder(6) | | | Glass(9) | | |
|---|---|---|---|---|---|---|---|---|---|---|---|---|---|---|---|
| | PRP (3) | CFS (2) | CON (4) | PRP (4) | CFS (8) | CON (11) | PRP (6) | CFS (8) | CON (8) | PRP (5) | CFS (5) | CON (4) | PRP (6) | CFS 6) | CON (7) |
| Naïve Bayes | 29.67 | 30.77 | 33.65 | 83.77 | 84.36 | 85.50 | 95.7 | 97.19 | 97.19 | 67.30 | 68.31 | 68.60 | 65.73 | 43.92 | 47.20 |
| SMO | 16.35 | 12.98 | 15.48 | 82.77 | 84.75 | 84.44 | 98.90 | 98.21 | 98.31 | 69.00 | 69.18 | 69.19 | 62.44 | 57.94 | 57.48 |
| KSTAR | 45.48 | 42.17 | 47.69 | 81.95 | 81.67 | 82.07 | 95.39 | 97.45 | 96.63 | 70.93 | 70.64 | 70.64 | 83.57 | 79.91 | 78.50 |
| Bagging | 50.0 | 45.07 | 50.77 | 80.40 | 81.11 | 81.48 | 94.86 | 94.94 | 94.94 | 72.22 | 70.64 | 71.22 | 76.53 | 73.83 | 71.50 |

| | 9 | | | | | | | | | | | | | | |
|---|---|---|---|---|---|---|---|---|---|---|---|---|---|---|---|
| **J48** | 42.65 | 38.08 | 41.61 | 82.31 | 81.11 | 82.89 | 96.0 | 93.82 | 94.94 | 68.90 | 68.31 | 69.48 | 72.30 | 68.69 | 64.20 |
| **PART** | 52.09 | 46.17 | 54.37 | 83.70 | 81.67 | 79.55 | 94.0 | 93.10 | 94.3 | 69.14 | 69.48 | 68.60 | 77.00 | 70.09 | 68.60 |
| **Average Accuracy** | 39.38 | 35.87 | 40.59 | 82.48 | 82.44 | 82.71 | 95.80 | 95.78 | 96.05 | 69.58 | 69.40 | 69.62 | 72.9 | 65.73 | 64.58 |

# 5. Conclusion

The relative indiscernibility relation introduces in the paper is an equivalence relation which induces a partition of equivalence classes for each attribute. Then, the degree of similarity is measured between two attributes based on their equivalence classes. Since, the target of the paper is to compute reduced attribute set for decision making, so application of equivalence classes for similarity measurement is the appropriate choice.

## Authors

Ms Shampa Sengupta is an Assistant Professor of Information Technology at MCKV Institute Of Engineering, Liluah, Howrah. She has received M.Tech degree in Information Technology from Bengal Engineering and Science University, Shibpur, Howrah. Since 2011, she has been working toward the PhD degree in Data Mining at Bengal Engineering and Science University, Shibpur, Howrah.

Dr.Asit Kr.Das is an Assistant Professor of Computer Science and Technology at Bengal Engineering and Science University, Shibpur, Howrah.He has received M.Tech degree in Computer Science and Engg from Calcutta University.He obtained PhD(Engg) degree from Bengal Engineering and Science University,Shibpur, Howrah. His research interests include Data Mining and Pattern Recognition and Rough Set Theory, Bio-informatics etc.